\documentclass{article}

\usepackage{PRIMEarxiv}

\usepackage[utf8]{inputenc} 
\usepackage[T1]{fontenc}    
\usepackage{url}            
\usepackage{microtype}      
\usepackage{fancyhdr}       

\usepackage{amsmath}
\usepackage{amssymb}
\usepackage{amsfonts}
\usepackage{booktabs}
\usepackage{graphicx}
\usepackage{subfig}         
\usepackage{float}
\usepackage[figuresright]{rotating} 

\usepackage[numbers]{natbib}  

\usepackage[bookmarks=false]{hyperref}
    \hypersetup{colorlinks,
      linkcolor=blue,
      citecolor=blue,
      urlcolor=blue}

\usepackage{doi}            

\title{HERBIOME: Automated pipeline for herbarium label digitization
\thanks{Accepted at the 30th International Conference on Knowledge-Based and
Intelligent Information \& Engineering Systems (KES 2026).
Source code available at: \url{https://github.com/IA-E-Col/HERBIOME}}}

\author{
  Hiba Abbad$^{\,a}$ \quad Hanane Ariouat$^{\,b}$ \quad Eva Perez Pimpare$^{\,d}$ \quad
  Nicolas Turenne$^{\,b,c}$ \quad Eric Chenin$^{\,b}$ \\
  Abderrazak Sebaa$^{\,a}$ \quad Edi Prifti$^{\,b,e}$ \quad
  Jean-Daniel Zucker$^{\,b,e}$ \quad Youcef Sklab$^{\,a,*}$ \\[6pt]
  \normalsize $^{a}$\'Ecole Sup\'erieure en Sciences et Technologies de l'Informatique et du Num\'erique (ESTIN), B\'eja\"ia, Algeria \\
  \normalsize $^{b}$IRD, Sorbonne Universit\'e, UMMISCO, Paris, France \\
  \normalsize $^{c}$INRAE, MathNum, France \\
  \normalsize $^{d}$Infrastructure R\'ecolnat, Direction g\'en\'erale d\'el\'egu\'ee aux collections, \\
  \normalsize \phantom{$^{d}$}Mus\'eum national d'histoire naturelle, Paris, France \\
  \normalsize $^{e}$Sorbonne Universit\'e, INSERM, Nutrition et Obesities; systemic approaches, NutriOmique, AP-HP, France \\[6pt]
  \normalsize $^{*}$Corresponding author: \texttt{youcef.sklab@ird.fr}
}

\begin{document}
\maketitle

\begin{abstract}
Digitized herbarium collections, now comprising over 100 million freely accessible specimen images, have become a critical resource for addressing fundamental questions in ecology and evolutionary biology. Yet the rich metadata encoded in herbarium labels (collector identities, geographic localities, collection dates, and ecological observations) remains largely inaccessible at scale, constraining both biodiversity informatics and the construction of specimen-specific image-text corpora for multimodal AI. We present \textbf{HERBIOME}, a modular end-to-end pipeline for automated herbarium label digitization, integrating YOLOv8-based component detection, CRAFT Hezar word-level text localization, fine-tuned TrOCR for recognition of mixed handwritten and printed text, and GPT-4o Mini for semantic metadata structuring into standardized fields. TrOCR was trained on a multi-source dataset combining general transcription corpora (CREMMA-AN, PictoCatalogs) with herbarium-specific data (R\'{e}ColNat), achieving a Character Error Rate of 4.05-4.10\%. End-to-end evaluation on 450 French herbarium specimens, using a dual-metric framework of Maximum Window Similarity (MWS: 0.614-0.618) and Semantic Metadata Accuracy (SMA: 0.440-0.445), reveals that hybrid training strategies improve semantic fidelity while random sampling maximizes surface similarity, with taxonomic fields remaining the principal bottleneck. By automating the extraction of structured metadata from complex, heterogeneous labels, HERBIOME reduces transcription burden, enables the construction of paired image-text datasets that faithfully capture specimen individuality, which is a prerequisite for next-generation multimodal biodiversity AI systems.
\end{abstract}



\section{Introduction}
\label{sec:introduction}

\begin{figure}[t]
    \centering
    \includegraphics[width=.5\linewidth]{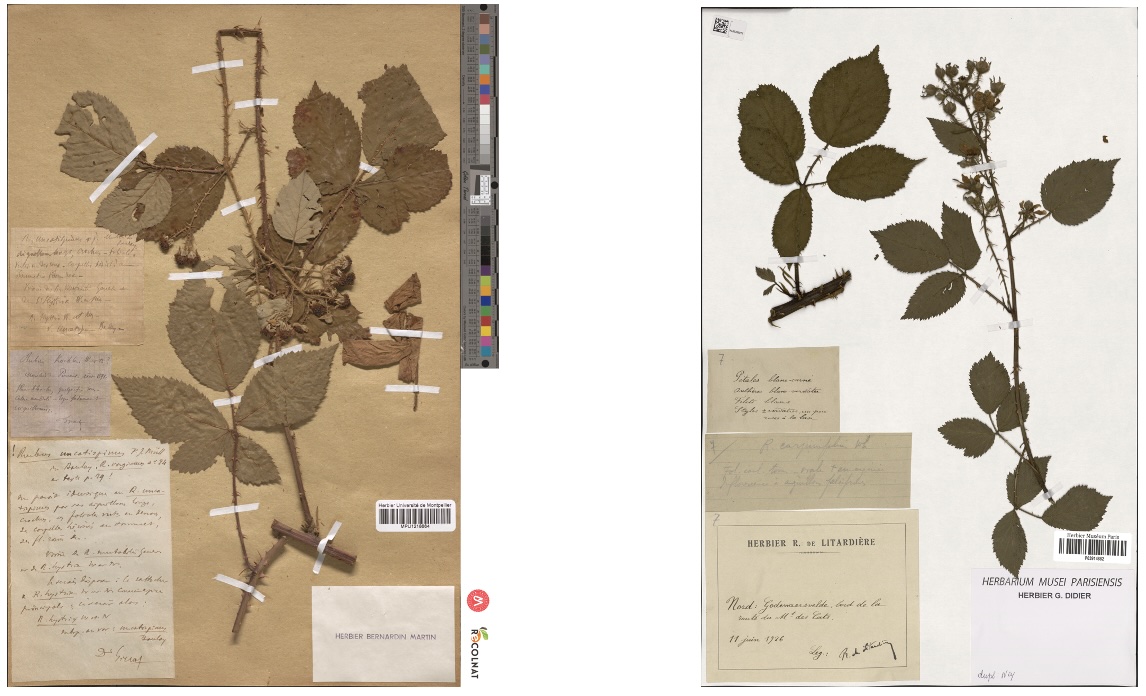}
    \caption{Two examples of digitized herbarium sheets from French collections
    (ReColNat). Handwritten annotations may record collection locality and date, ecological observations, regional habitat notes, and ethnobotanical uses, as well as subsequent determinations added by visiting botanists over decades. Two specimens of the same species may thus differ substantially in both morphology and documentary content, making each sheet an irreplaceable record whose full scientific value can only be unlocked through systematic label digitization.}
    \label{fig:herbarium_examples}
\end{figure}

The world's herbaria collectively hold more than 400 million preserved plant specimens, representing centuries of botanical exploration across all continents \cite{james2018herbarium}. Large-scale digitization initiatives have made over 100 million high-resolution specimen images freely accessible online \cite{sklab2025plantsam}, opening unprecedented opportunities for data-driven botanical research. Digitized collections are increasingly deployed to address fundamental questions in ecology and evolutionary biology: tracking phenological shifts across continental extents and multi-century timescales \cite{ahlstrand2025phenology}, reconstructing leaf economic traits correlated with climate gradients across thousands of species \cite{vasconcelos2025lma}, and enabling automated species identification at near-expert accuracy \cite{shirai2022identification}. Yet the scientific exploitation of these collections remains critically constrained by a persistent metadata gap. Each herbarium sheet carries one or more labels recording taxonomic identifications, collection dates, geographic localities, collector names, habitat descriptions, and determination history\footnote{A specimen-specific documentary record that is unique and irreplaceable.} \cite{FiggViruel2024, Engledow2022}. As illustrated in Figure~\ref{fig:herbarium_examples}, a herbarium sheet accumulates multiple layers of documentary content over time: beyond the pressed plant itself, labels, stamps, barcodes, and color charts encode institutional provenance, while handwritten annotations added by botanists across decades may record collection locality and date, phenological stage, ecological observations, habitat specifics, and occasionally ethnobotanical uses of the collected plant. Labels thus serve as the primary key to understanding each specimen's provenance, ecological context, and scientific history. Despite their scientific value, the vast majority of these labels remain untranscribed or only partially digitized, leaving an immense reservoir of structured and contextual knowledge locked in analog form.

Manual transcription is labor-intensive, error-prone, and impossible to sustain at the scale of millions of sheets \cite{Engledow2022}. Existing OCR tools, including Tesseract \cite{smith2007} and EasyOCR \cite{jaided2020easyocr}, partially address printed text extraction but fail on handwritten, degraded, or multilingual content typical of historical herbarium labels \cite{drinkwater2014ocr, Hussein2021}. Most automated pipelines focus on a limited subset of metadata fields and continue to require expert validation for heterogeneous or damaged labels, severely limiting their scalability across institutional collections \cite{turnbull2025hespi, guralnick2024humans}. To address these limitations, we present \textbf{HERBIOME}, a modular end-to-end pipeline for the automated digitization of herbarium labels. HERBIOME integrates four complementary stages: i) YOLOv8-based component detection to isolate text-bearing elements from full sheet scans; ii) CRAFT Hezar \cite{hezarai_craft_2024} word-level text localization with reading-order restoration; iii) fine-tuned TrOCR \cite{li2021trocr} for recognition of mixed handwritten and printed text; and iv) GPT-4o Mini for semantic metadata structuring into standardized fields. The pipeline is designed for the specific challenges of French herbarium collections (complex label layouts, cursive historical handwriting, Latin nomenclature, and mixed typography) while remaining modular enough to support component-level substitution as the field advances. The main contributions of this paper are:
\begin{itemize}
	\item A modular four-stage HERBIOME pipeline integrating component detection, word-level text localization, transformer-based OCR, and LLM-driven metadata structuring in a unified, reproducible end-to-end workflow.
	\item A multi-source training strategy for TrOCR combining general transcription corpora (CREMMA-AN, PictoCatalogs) with herbarium-specific data (R\'eColNat), achieving a Character Error Rate of 4.05-4.10\% on mixed handwritten and printed French text.\footnote{\url{https://www.recolnat.fr}}
	\item A dual-metric evaluation framework combining Maximum Window Similarity (MWS) and Semantic Metadata Accuracy (SMA), enabling comprehensive assessment of both surface-level transcription quality and semantic correctness of structured metadata fields.
	\item An end-to-end evaluation on 450 real French herbarium specimens from the IRIS dataset\footnote{A dataset produced and annotated by domain experts and citizen scientists through \textit{Les Herbonautes}, a collaborative transcription platform {\url{https://www.lesherbonautes.mnhn.fr}}.}, demonstrating practical performance, identifying field-specific bottlenecks, and discussing implications for large-scale biodiversity platforms.
\end{itemize}

\section{Related work}
\label{sec:related-work}

The digitization of herbarium collections (see Figure \ref{fig:pipeline}) has catalyzed a growing ecosystem of AI-driven tasks, spanning automated species identification \cite{shirai2022identification}, morphological trait extraction \cite{vasconcelos2025lma}, and large-scale phenological research linking specimens to climate change \cite{ahlstrand2025phenology, tessler2025herbarium, guo2025aireview}. Within this ecosystem, automated label digitization is a structurally critical step: it recovers the specimen-specific metadata that enriches visual records and connects individual sheets to global biodiversity platforms. This process generally follows a three-stage workflow (\textbf{detection}, \textbf{optical character recognition (OCR)}, and \textbf{post-OCR structuring}) each posing specific challenges due to heterogeneous layouts, mixed handwritten and printed content, multilingual text, and varying degrees of degradation.

\paragraph{\textbf{Label and component detection}}

Accurate localization of labels, barcodes, stamps, and other components (see Figure \ref{fig:pipeline}) is essential for downstream OCR performance. General-purpose detectors such as MaX-DeepLab, DBNet, and TextBPN++ and text-specific models such as SwinTextSpotter and CRAFT perform well on structured documents \cite{Long_2022_CVPR, olejniczak2023text, huang2022swintextspotter, baek2019craft}, but are unlikely to generalize well to herbarium sheets, which present overlapping plant material, irregular layouts, and low-resolution scans. Herbarium-specific YOLO-based approaches address these challenges more effectively: YOLOv5 achieved strong mAP and F1 scores within single collections, though cross-institutional generalization remains limited \cite{thompson2023identification, dillen2019benchmark}, and the HESPI pipeline's two-stage YOLOv8 system reaches up to 98.5\% F1 on institutional labels \cite{turnbull2025hespi}. Beyond label localization, dedicated models have demonstrated precise detection of plant organs \cite{ariouat2024yolov7} and non-botanical sheet elements — stamps, barcodes, envelopes, and color charts \cite{sklab2025nonplant} — confirming that the full visual content of herbarium sheets can be systematically parsed. Despite these advances, field-level detection alone does not capture complete metadata, motivating HERBIOME's modular design combining coarse component detection with fine-grained text localization.

\paragraph{\textbf{Optical character recognition (OCR)}}

OCR for herbarium labels is complicated by the coexistence of printed and handwritten text, historical degradation, and multilingual content. CNN-based pipelines (CNN-BiLSTM-CTC) achieve CER of 10-15\% on standard datasets \cite{swaileh2018unified, olejniczak2023text}, while Tesseract-cloud OCR(Google Vision) ensembles improve mixed-content performance at the cost of pre-classification and human verification \cite{owen2020nlpworkflow, guralnick2024humans}. Transformer-based models, including TrOCR, offer robust recognition of irregular handwriting and long-range dependencies, though at higher computational cost \cite{marttila2024htr, Rang_2024_CVPR}; hybrid CNN-Transformer architectures such as SwinTextSpotter and HTRVT balance flexibility with efficiency \cite{huang2022swintextspotter, li2024htrvt}. In herbarium-specific pipelines, HESPI integrates separate engines for printed and handwritten text, achieving strong results on curated institutional datasets but showing reduced generalization across more diverse collections \cite{turnbull2025hespi, dillen2019benchmark}. These observations confirm that robust herbarium OCR benefits from hybrid architectures capable of handling heterogeneous text without extensive pre-classification.

\paragraph{\textbf{Post-OCR structuring}}

After recognition, extracted text must be structured into metadata aligned with biodiversity standards such as Darwin Core. NER-based approaches reliably identify taxonomic names, collection dates, localities, and collectors at low computational cost \cite{owen2020nlpworkflow, takano2024automated}, while LLMs such as GPT-4o handle noisy OCR outputs, multilingual content, and variable label formats \cite{weaver2023herbarium, turnbull2025hespi}. Hybrid NER-LLM pipelines improve both accuracy and semantic coverage, though human verification remains necessary for complex labels \cite{guralnick2025ensemble}. Fully automated extraction of complete metadata across heterogeneous collections remains an open challenge.

\paragraph{\textbf{From label digitization to multimodal herbarium AI}}

The three-stage pipeline is one layer of a broader analytical stack. Whole-specimen segmentation pipelines such as PlantSAM \cite{sklab2025plantsam}, combined with interpretability-driven studies of trait classification \cite{ariouat2025interpretable}, demonstrate that isolating plant foreground substantially improves downstream morphological analysis. Attention-guided and multimodal architectures further consolidate plant-centric feature learning by integrating 2D appearance with geometric or segmentation-derived representations \cite{sedrat2026atvit, sklab2025simnet}. Recovering structured label metadata from these same specimens unlocks a complementary dimension: it enables the construction of paired image-text corpora that ground visual representations in specimen-specific ecological and historical context, a prerequisite for next-generation multimodal biodiversity AI \cite{tessler2025herbarium, guo2025aireview}. These limitations and opportunities motivate HERBIOME's modular design, which integrates two-stage detection, hybrid OCR, and semantic post-processing specifically tailored for French herbarium specimens.

\section{Proposed methodology: HERBIOME pipeline}
\label{sec:method}

\begin{figure}[t]
	\centering
	\includegraphics[width=0.7\textwidth]{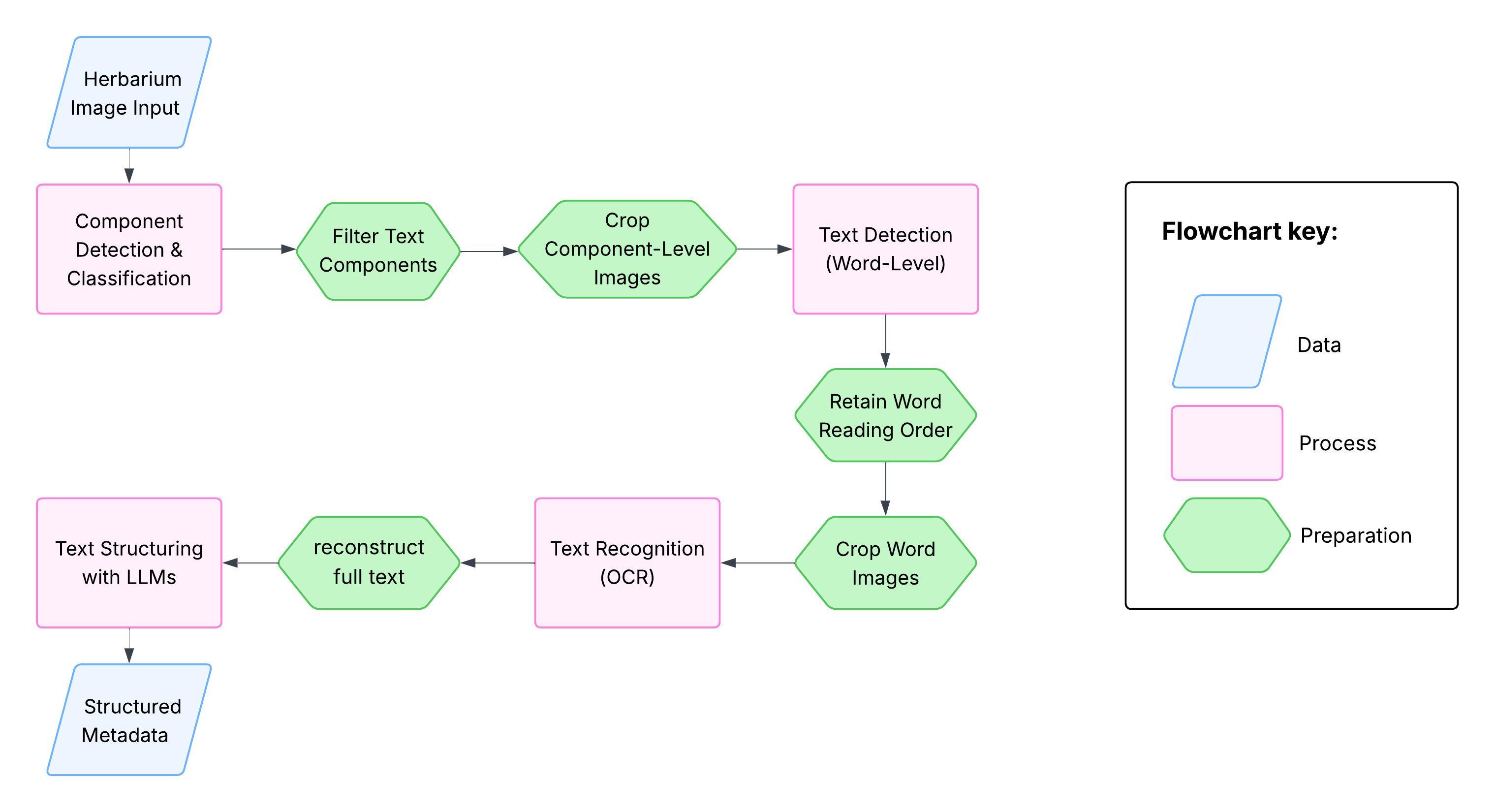}
	\caption{Overview of the \textbf{HERBIOME} pipeline: from raw herbarium
		sheet images to structured metadata.}
	\label{fig:herbiome-pipeline}
\end{figure}

The HERBIOME pipeline transforms raw herbarium sheet images into structured, machine-readable metadata through four sequential stages, illustrated in Figure~\ref{fig:herbiome-pipeline}.

\subsection{Stage 1: Component detection and classification}
\label{sec:component_detection}
The first stage identifies and isolates text-bearing elements from full sheet scans. 
HERBIOME employs the HESPI YOLOv8-x model \cite{turnbull2025hespi}, which detects 
eleven component types: primary specimen labels, handwritten data outside labels, 
annotation labels, herbarium stamps, swing tags, numbers outside primary labels, small, 
medium, and full database labels, color targets, and measurement scales. Primary label 
detection achieves F1 = 98.5\%, with F1 scores between 95\% and 97\% for other types. 
Only components with confidence $s \geq 0.7$ belonging to text-related classes (primary, 
handwritten, annotation, and database labels) are retained and cropped for subsequent 
processing. Each retained component is represented as a tuple $(c, b, s)$, where $c$ is 
the predicted class, $b = (x_1, y_1, x_2, y_2)$ the bounding box, and $s \in [0,1]$ 
the confidence score. Figure~\ref{fig:pipeline} illustrates the diversity 
of detected components. The threshold of 0.7 was selected as a quality filter to retain 
only high-confidence detections, minimizing false positives passed to downstream OCR 
stages. Since the HESPI model was adopted as-is from its original validated 
implementation, retraining or threshold optimization was not required, and false 
negatives at this threshold remain negligible in practice.

\subsection{Stage 2: Word-level text detection}
\label{sec:word_level_detection}

\begin{figure}[t]
	\centering
	\begin{minipage}{0.55\linewidth}
		\centering
		\subfloat[Raw CRAFT output (incorrect reading flow).]{%
			\includegraphics[width=0.38\linewidth]{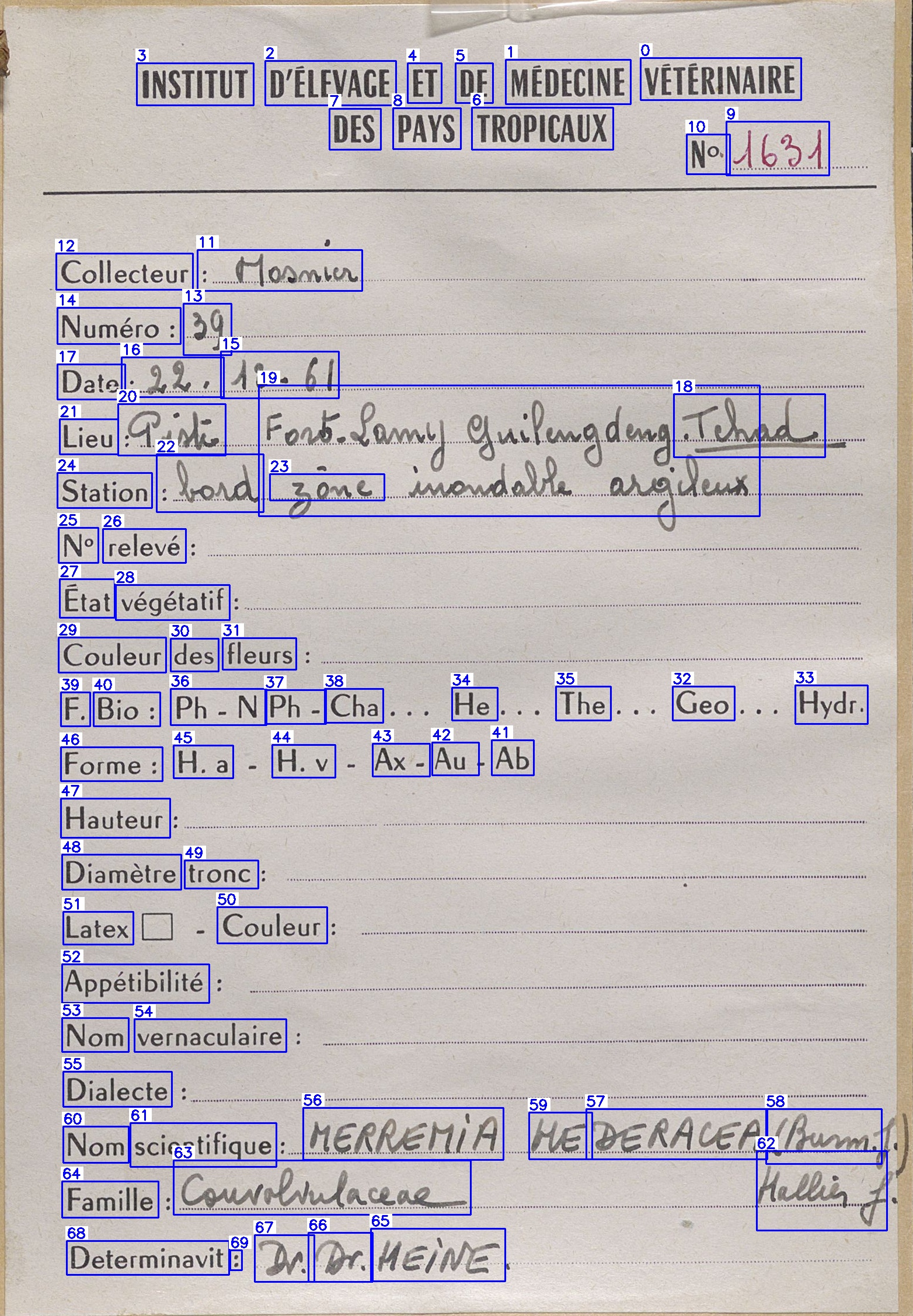}%
			\label{fig:craft_order}}\hfill
		\subfloat[DBSCAN-corrected ordering (natural reading flow).]{%
			\includegraphics[width=0.38\linewidth]{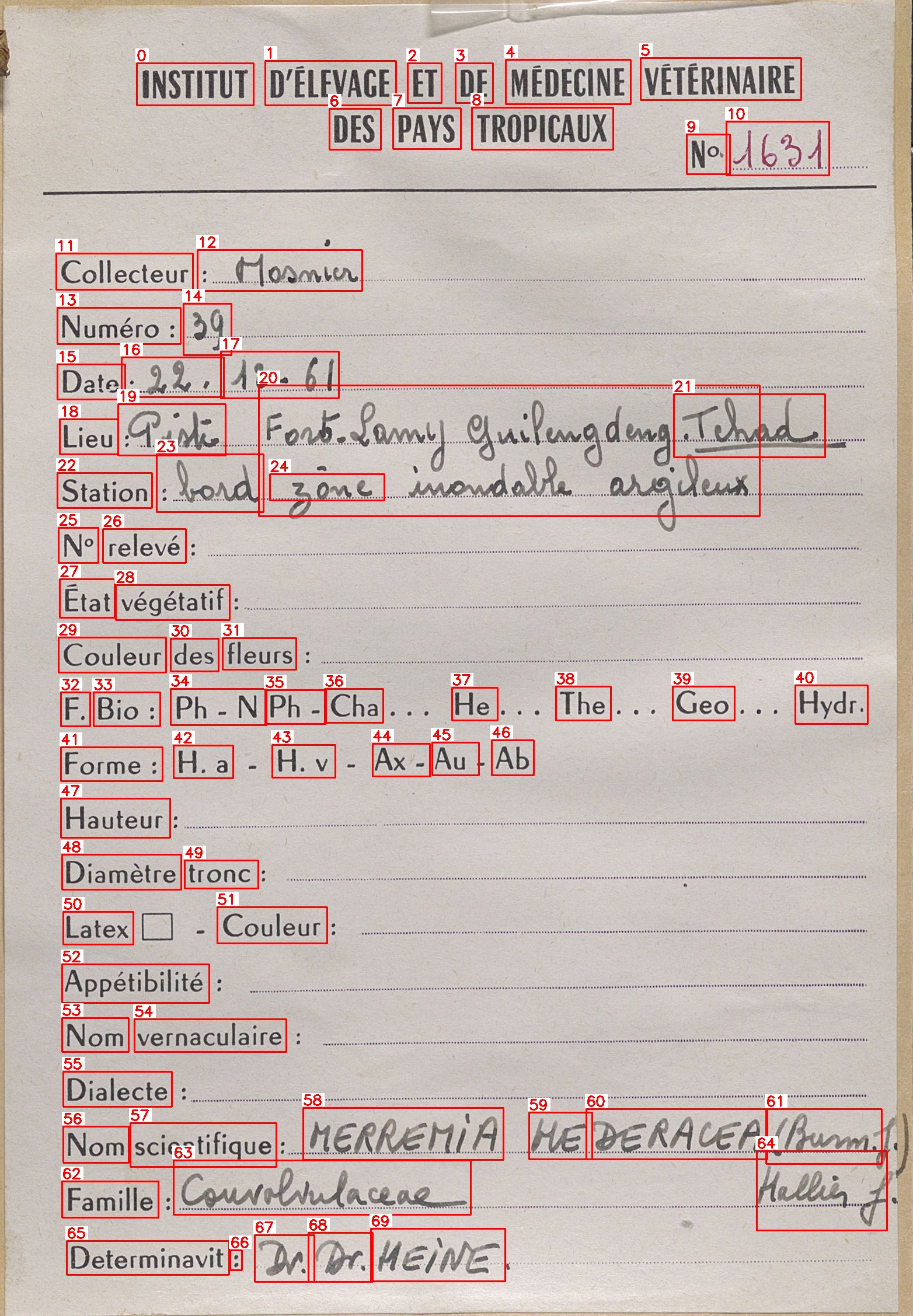}%
			\label{fig:dbscan_order}}
	\end{minipage}
	\caption{Word ordering: raw CRAFT output vs.\ DBSCAN-corrected result.}
	\label{fig:order_comparison}
\end{figure}

Individual words are detected within each cropped component using the \textbf{CRAFT Hezar} model \cite{hezarai_craft_2024}, which localizes character regions and their affinities, handling curved, skewed, and handwritten text. Each word region is defined by a bounding box $b = (x_1, y_1, x_2, y_2)$, center $c = \!\left(\frac{x_1+x_2}{2},\,\frac{y_1+y_2}{2}\right)$, and height
$h = y_2 - y_1$. Since CRAFT does not preserve reading order, a post-processing step based on \textbf{DBSCAN clustering} groups word centers vertically with $\varepsilon = 0.6\cdot\tilde{h}$, where $\tilde{h}$ is the median word height: words $c_i$ and $c_j$ are assigned to the same line if $|y_i - y_j| \leq \varepsilon$. Lines are then sorted top-to-bottom and words within each line left-to-right, restoring natural reading flow for irregular or curved layouts. Figure~\ref{fig:order_comparison} illustrates the reordering effect.

\subsection{Stage 3: Text recognition}
\label{sec:stage3_text_recognition}

Ordered word crops are transcribed using a fine-tuned \textbf{TrOCR} model \cite{li2021trocr}, an encoder-decoder Transformer that maps image patches directly to character sequences via cross-attention, bypassing the CTC decoding of conventional OCR pipelines. Given a crop $I \in \mathbb{R}^{H \times W \times 3}$, the visual encoder $E_\theta$ produces feature vectors $z = E_\theta(I) \in \mathbb{R}^{T \times d}$; the decoder $D_\phi$ then autoregressively generates $\hat{y} = \{\hat{y}_1, \dots, \hat{y}_n\}$ with probability $P(\hat{y}\mid I) = \prod_{t=1}^{n} P(\hat{y}_t \mid \hat{y}_{<t}, z)$. Beam search is applied at inference. This architecture provides resilience to handwriting variability, degraded ink, and the mixed typography characteristic of historical herbarium labels.

\subsection{Stage 4: Metadata structuring}
\label{sec:metadata_structuring}

\begin{figure}[t]
	\centering
	\includegraphics[height=0.6\textheight, keepaspectratio]{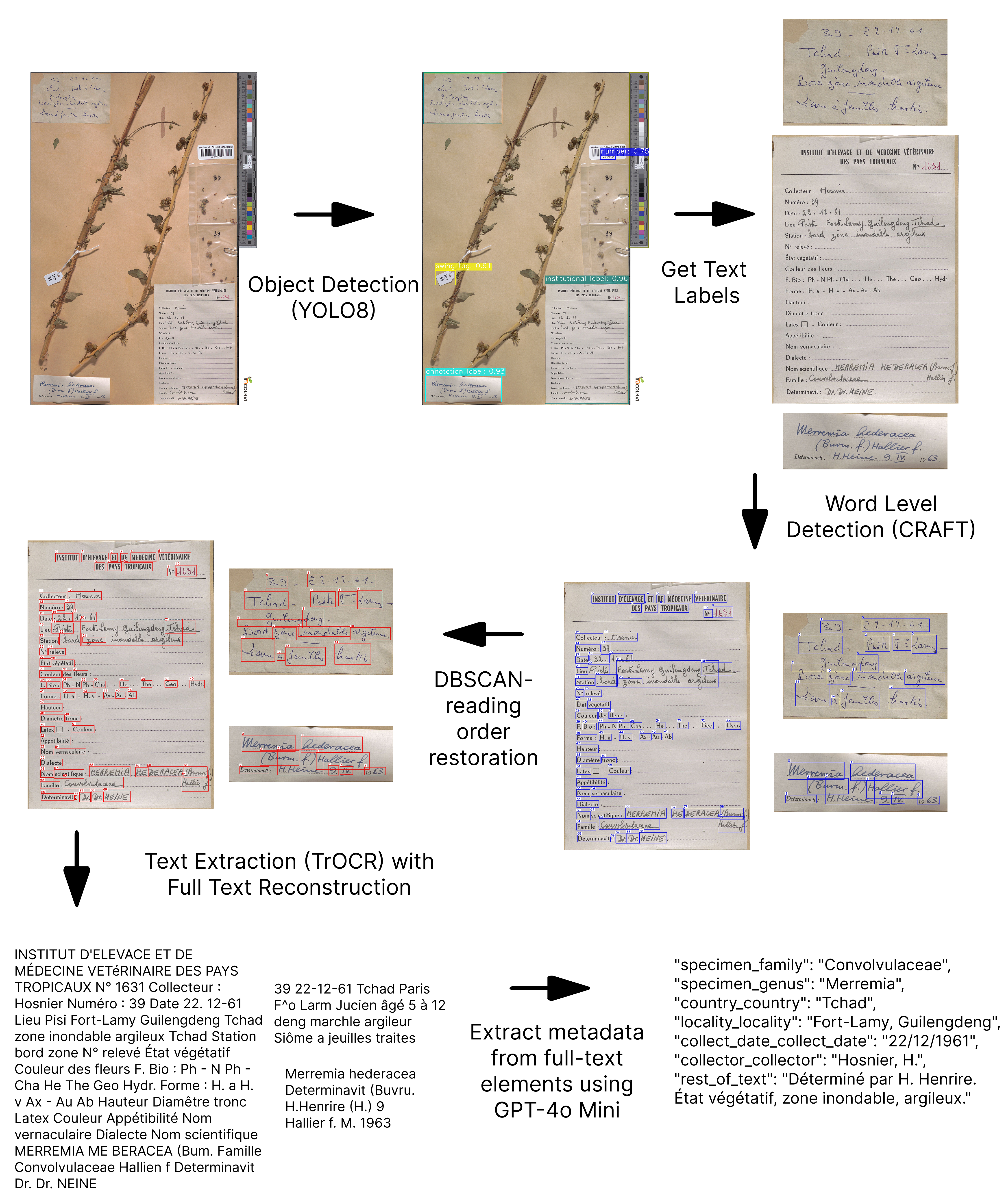}
	\caption{End-to-end HERBIOME example: component detection, word-level
		localization, TrOCR transcription, and GPT-4o Mini metadata structuring.}
	\label{fig:pipeline}
\end{figure}

OCR output is transformed into structured metadata using \textbf{GPT-4o Mini}, prompted to extract seven fields from potentially noisy, misordered text: \texttt{specimen\_family}, \texttt{specimen\_genus}, \texttt{country}, \texttt{locality}, \texttt{collect\_date}, \texttt{collectors}, and \texttt{rest\_of\_text}. Domain-specific prompts enforce JSON output, correct common OCR errors, and prevent field duplication. Post-processing rules ensure cross-collection consistency:
\begin{itemize}
	\item Dates are normalized to \texttt{DD/MM/YYYY\,-\,DD/MM/YYYY},
	\texttt{MM/YYYY}, or \texttt{YYYY}.
	\item Country names are standardized to official forms.
	\item Localities are stripped of extraneous taxonomic tokens or collection codes.
	\item Missing fields are retained as empty strings rather than inferred.
\end{itemize}
Figure~\ref{fig:pipeline} presents a complete end-to-end example, from a raw herbarium scan to the final structured JSON output.

\section{Case study and experimental evaluation}
\label{sec:case_study}

Four datasets were used for training and evaluation. \textbf{PictoCatalogs} \cite{htrcatalogs_dataset} provides 19\textsuperscript{th}-20\textsuperscript{th}-century French printed catalogs, supporting OCR fine-tuning on diverse typography and Latin scientific names. \textbf{CREMMA-AN} \cite{cremma_testaments_poilus} contributes 146 handwritten French wills (1898-1918), reinforcing robustness to cursive and degraded scripts. \textbf{ReColNat Word-Level} supplies $\approx$1,100 annotated word-level images for herbarium-specific fine-tuning. \textbf{IRIS}, compiled from RECOLNAT, is reserved exclusively for end-to-end pipeline evaluation and provides ground truth across all four stages. Its annotations were produced by domain experts and citizen scientists through the \textit{Les Herbonautes} collaborative transcription
platform\footnote{\url{https://www.lesherbonautes.mnhn.fr}}, which coordinates voluntary contributions to the digitization of French natural history collections. For OCR training, 3,330 CREMMA-AN line images and 5,487 PictoCatalogs samples were merged with 1,124 ReColNat word-level images and split 80\% for train and 20\% for validation.

\subsection{Experimental setup}
\label{sec:experimental_setup}

Three training strategies were evaluated: \textbf{Random Sampling} and \textbf{Hybrid Sampling} apply their sampling strategy at the word level on the merged dataset; \textbf{Two-Stage Training} uses line-level images first, then fine-tunes on word-level data. Preprocessing was applied uniformly:
\begin{itemize}
	\item \textit{Image}: conversion to RGB, fixed contrast and sharpening, resize to
	$384\times384$ px with aspect-ratio padding.
	\item \textit{Text}: tokenization with TrOCR's tokenizer (max length 64, BOS token
	prepended); corrupted samples replaced with placeholders.
	\item \textit{Augmentation} (20\%-80\%): affine and perspective distortions,
	brightness/CLAHE adjustments, Gaussian and ISO noise injection.
\end{itemize}

Training used FP16 mixed precision, effective batch size 256 (8 per device $\times$ 32 gradient accumulation steps), polynomial scheduler (base $1.5\times10^{-5}$, 20\% warm-up), weight decay 0.008, gradient clipping 0.3, and label smoothing $\alpha=0.08$. Hybrid Sampling additionally enforced a 70/30 line-to-word curriculum ratio per batch. Early stopping triggered after 20 evaluations without CER improvement; beam search was applied at inference.

\paragraph{Evaluation metrics}
OCR accuracy was measured with Character Error Rate $\text{CER} = (S+D+I)/N$, Word Error Rate $\text{WER} = (S_w+D_w+I_w)/N_w$, and Exact Match (EM). End-to-end metadata quality was assessed with two complementary metrics. \textbf{Semantic Metadata Accuracy (SMA)} scores each field $f$ in record $i$ as $S^f_i \in \{1.0, 0.5, 0.0\}$ (semantically equivalent, partially correct, or incorrect) and aggregates as $\text{SMA} = \sum_{i,f} S^f_i / (N \times F)$. \textbf{Maximum Window Similarity (MWS)} evaluates surface-level robustness: $\text{MWS}^f_i = \max_{w \in W^f_i} \operatorname{fuzzy\_partial}(\operatorname{norm}(\hat{p}^f_i),
\operatorname{norm}(w))$, where $W^f_i$ is the set of ground-truth substring windows. CER, WER, and EM capture transcription fidelity; SMA and MWS capture semantic and surface correctness of structured metadata respectively.

\subsection{Results}
\label{sec:results}

\begin{figure}[t]
	\centering
	\subfloat[Random Sampling]{%
		\includegraphics[width=0.48\linewidth]{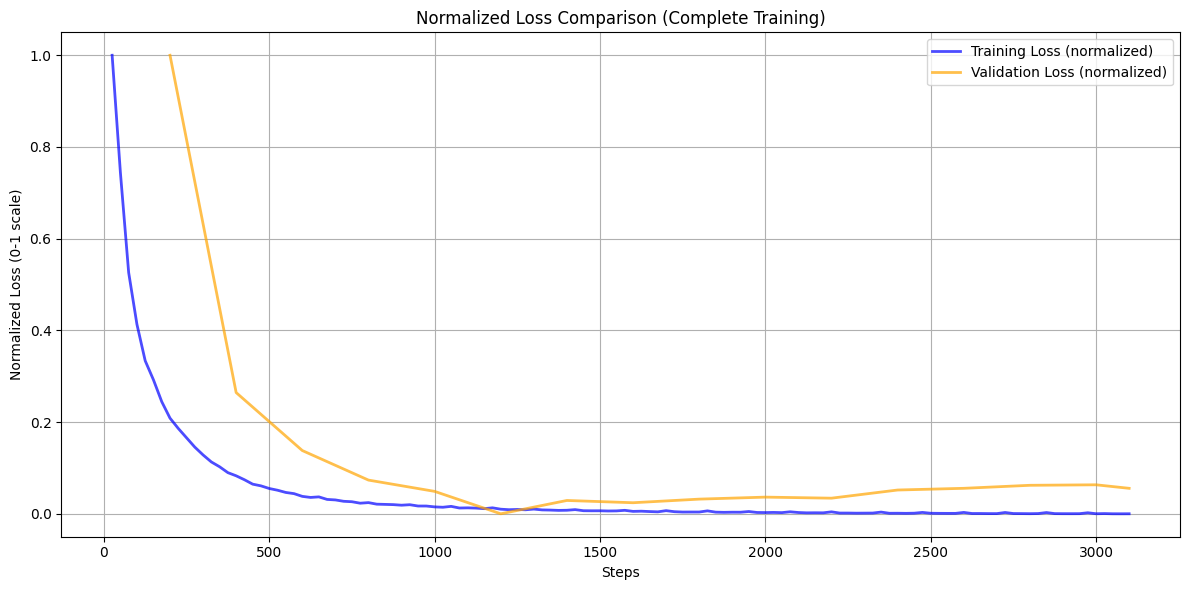}}\hfill
	\subfloat[Hybrid Sampling]{%
		\includegraphics[width=0.48\linewidth]{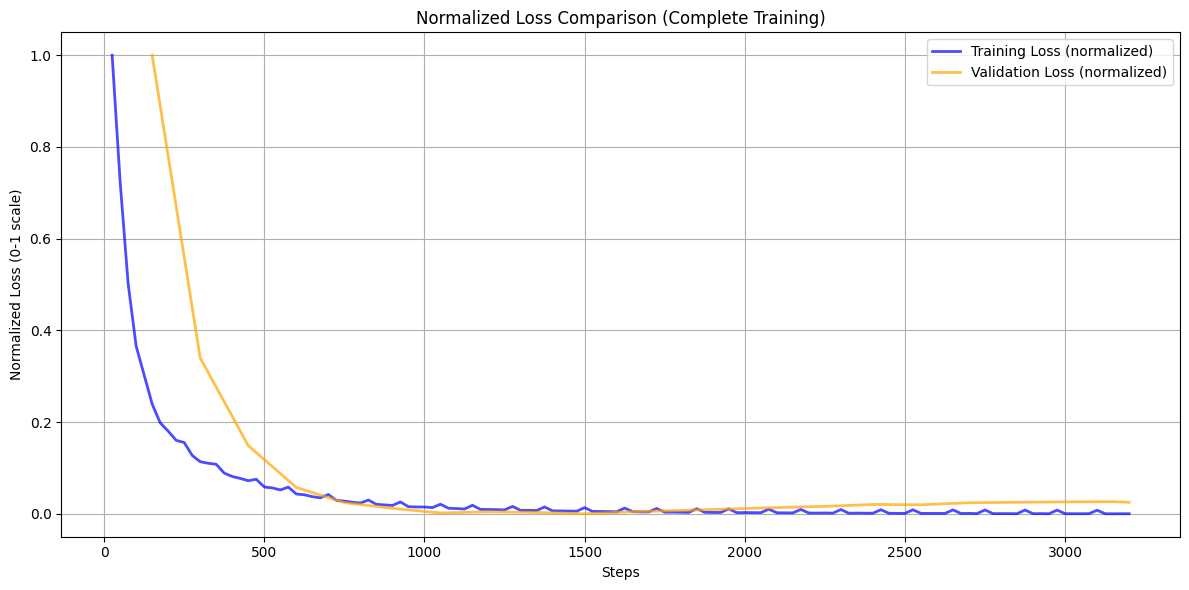}}
	\caption{Normalized training and validation loss dynamics for Random and Hybrid
		sampling strategies.}
	\label{fig:training_dynamics}
\end{figure}

OCR results are reported in Table~\ref{tab:pipeline_evaluation}(a). Random Sampling achieves the lowest CER (4.05\%) and highest EM (69.3\%), while Hybrid Sampling exhibits more stable training dynamics (Figure~\ref{fig:training_dynamics}) at a marginal cost in raw accuracy. Two-Stage Training generalizes poorly on mixed scripts (CER: 5.64\%).

\begin{table}[t]
	\caption{Pipeline evaluation: (a) OCR performance, (b) overall MWS, and (c) semantic
		metadata accuracy.}\label{tab:pipeline_evaluation}
	\centering
	\begin{tabular*}{\hsize}{@{\extracolsep{\fill}}lccc@{}}
		\toprule
		\multicolumn{4}{@{}l}{\textit{(a) OCR evaluation results}} \\
		\midrule
		\textbf{Configuration} & \textbf{CER (\%)} & \textbf{WER (\%)} & \textbf{Exact Match (\%)} \\
		\midrule
		Random Sampling     & \textbf{4.05} & \textbf{11.2} & \textbf{69.3} \\
		Hybrid Sampling     & 4.10          & 11.4          & 67.3 \\
		Two-Stage Training  & 5.64          & 14.1          & 63.8 \\
		\midrule
		\multicolumn{4}{@{}l}{\textit{(b) Overall MWS (IRIS, 450 specimens)}} \\
		\midrule
		\textbf{Model} & \textbf{Mean MWS} & \textbf{Specimens} & \textbf{Threshold} \\
		\midrule
		Random Sampling & \textbf{0.618} & 450 & 0.7 \\
		Hybrid Sampling & 0.614          & 450 & 0.7 \\
		\midrule
		\multicolumn{4}{@{}l}{\textit{(c) Semantic Metadata Accuracy (450 specimens)}} \\
		\midrule
		\textbf{Model} & \textbf{Average SMA} & \textbf{Score Range} & \textbf{Semantic Quality} \\
		\midrule
		Hybrid Sampling & \textbf{0.445} & 0.0-1.0 & \textbf{Best} \\
		Random Sampling & 0.440          & 0.0-1.0 & Strong \\
		\bottomrule
	\end{tabular*}
\end{table}

End-to-end results on 450 IRIS specimens are reported in Tables~\ref{tab:pipeline_evaluation}(b,c) and~\ref{tab:field_performance}. Collection dates are the most reliably extracted field (MWS $>0.81$, accuracy $>0.66$), benefiting from standardized formats and preprocessing normalization. Taxonomic fields (family, genus) are the most error-prone (accuracy $<0.55$), reflecting the sensitivity of Latin nomenclature to minor OCR errors and motivating continued human-machine collaboration for specialized fields. Hybrid Sampling outperforms Random on semantic fidelity (SMA: 0.445 vs.\ 0.440) despite similar surface scores, confirming that curriculum-based strategies better handle format and style variation across heterogeneous
labels. The divergence between MWS and SMA rankings illustrates the complementary value of the dual-metric framework: surface similarity alone would obscure the semantic robustness essential for reliable integration into biodiversity platforms such as GBIF.

\begin{table}[t]
    \caption{Field-level performance: accuracy (similarity $\geq 0.9$)
    and mean MWS per field.}\label{tab:field_performance}
    \centering
    \begin{tabular*}{\hsize}{@{\extracolsep{\fill}}lccc||ccc@{}}
        \toprule
        & \multicolumn{3}{c}{\textit{\textit{(a) Accuracy (similarity $\geq 0.9$)}}}
        & \multicolumn{3}{c}{\textit{\textit{(b) Mean MWS per field}}} \\
        \midrule
        \textbf{Field}
            & \textbf{Random} & \textbf{Hybrid} & \textbf{Best}
            & \textbf{Random} & \textbf{Hybrid} & \textbf{Best} \\
        \midrule
        specimen\_family
            & \textbf{0.407} & 0.380          & \textbf{Random}
            & \textbf{0.450} & 0.427          & \textbf{Random} \\
        specimen\_genus
            & \textbf{0.541} & 0.495          & \textbf{Random}
            & \textbf{0.634} & 0.602          & \textbf{Random} \\
        country
            & 0.495          & \textbf{0.500} & \textbf{Hybrid}
            & 0.622          & \textbf{0.628} & \textbf{Hybrid} \\
        collect\_date
            & 0.664          & \textbf{0.716} & \textbf{Hybrid}
            & 0.815          & \textbf{0.835} & \textbf{Hybrid} \\
        locality
            & 0.466          & \textbf{0.468} & \textbf{Hybrid}
            & \textbf{0.648} & \textbf{0.648} & Tie \\
        collector
            & \textbf{0.468} & \textbf{0.468} & Tie
            & 0.541          & \textbf{0.543} & \textbf{Hybrid} \\
        \bottomrule
    \end{tabular*}
\end{table}

\paragraph{Comparison with Prior Systems}
A direct, metric-for-metric comparison with prior end-to-end herbarium 
digitization pipelines is not possible here because no shared benchmark 
exists: prior systems are evaluated on disjoint corpora, target a different 
and typically smaller set of fields, and report metrics that are not 
defined over the same objects as ours. HESPI~\cite{turnbull2025hespi}, the 
most closely related pipeline, trains a dedicated label-field detector on 
3{,}642 annotated institutional label images to visually localize and 
isolate a fixed set of fields (family, genus, species, collector), then 
applies separate OCR engines for printed and handwritten text with LLM-based 
post-correction. HERBIOME, by contrast, reconstructs and transcribes full 
label text across seven fields without any pre-trained field detector, and 
must handle unstructured handwritten content where field boundaries cannot 
be visually delimited. The two systems therefore measure different things: 
HESPI scores text similarity on visually isolated fields, whereas our MWS 
and SMA score, respectively, surface similarity and semantic correctness of 
metadata reconstructed from full-label text.

To bound the gap qualitatively despite this mismatch, we note that MWS is a 
normalized fuzzy partial-similarity score in $[0,1]$ and is thus loosely 
commensurable with HESPI's reported text-similarity range. HESPI attains 
91.3-95.9\% text similarity on clean institutional labels (MELU) but drops 
to 71.8\% on the more heterogeneous DILLEN collection. HERBIOME reaches a 
mean MWS of 0.614-0.618 on the heterogeneous, mixed handwritten/printed 
French historical specimens of IRIS. These figures are not directly 
comparable and should be read only as an order-of-magnitude positioning: 
HERBIOME operates closer to HESPI's harder, cross-collection regime than to 
its clean-label regime, while attempting a strictly broader task (full-label 
transcription over seven fields, no field detector, historical cursive 
French). HESPI's own 91.3\,$\rightarrow$\,71.8\% drop across collections 
indicates that field-detection pipelines are sensitive to label 
heterogeneity (the cregime HERBIOME is designed for) though we cannot 
claim from these data alone that HERBIOME escapes that sensitivity. A shared 
evaluation of both pipelines on a common heterogeneous corpus, under MWS/SMA, 
remains necessary for a conclusive comparison and is left to future work.

\section{Conclusion}
\label{sec:conclusion}

HERBIOME demonstrates that modular integration of detection, transformer-based OCR, and LLM-driven structuring can automate the extraction of structured metadata from complex, heterogeneous herbarium labels at competitive accuracy. The dual-metric evaluation framework (combining MWS for surface robustness and SMA for semantic fidelity) reveals a systematic trade-off between training strategies that a single metric would conceal, and provides a reusable benchmark design for end-to-end digitization pipelines. Taxonomic fields remain the principal bottleneck, driven by Latin nomenclature sensitivity to minor recognition errors, and will require targeted domain adaptation or hybrid human-machine validation to reach the accuracy thresholds demanded by platforms such as GBIF \cite{gbif}. 

Current limitations include the relatively small herbarium-specific training corpus and exclusive focus on French collections. Future work should extend to multilingual and multi-institutional corpora and incorporate stage-specific benchmarking. Beyond
metadata recovery, the structured label outputs produced by HERBIOME directly enable the construction of paired image-text datasets that capture specimen individuality, a resource that is currently absent at scale and that constitutes a prerequisite for
training next-generation multimodal biodiversity AI systems capable of jointly reasoning over visual morphology and ecological provenance.

\section*{Acknowledgements}
The authors gratefully acknowledge all citizen scientists who contributed to the annotation of herbarium specimen labels through the \textit{Les Herbonautes} platform (\url{https://www.lesherbonautes.mnhn.fr}), coordinated by the Muséum national d'Histoire naturelle (MNHN), Paris.


\bibliographystyle{unsrtnatdoi}
\bibliography{references}

\end{document}